%% file: main.tex
\documentclass{article}

\usepackage{iclr2025_conference,times}

\usepackage[utf8]{inputenc} 
\usepackage[T1]{fontenc}    
\usepackage{hyperref}       
\usepackage{url}            
\usepackage{booktabs}       
\usepackage{amsfonts}       
\usepackage{nicefrac}       
\usepackage{microtype}      
\usepackage{xcolor}         
\usepackage{makecell}
\usepackage{amsmath}
\usepackage{natbib}

\usepackage{longtable}     
\usepackage{caption}   
\usepackage{url}         
\usepackage{array,subcaption,multirow,graphicx}
\usepackage{listings,todonotes,csquotes}
\usepackage{caption}
\definecolor{codegreen}{rgb}{0,0.6,0}
\definecolor{codegray}{rgb}{0.5,0.5,0.5}
\definecolor{codepurple}{rgb}{0.58,0,0.82}
\definecolor{backcolour}{rgb}{0.95,0.95,0.92}

\lstdefinestyle{mystyle}{
    backgroundcolor=\color{backcolour},   
    commentstyle=\color{codegreen},
    keywordstyle=\color{magenta},
    numberstyle=\tiny\color{codegray},
    stringstyle=\color{codepurple},
    basicstyle=\ttfamily\footnotesize,
    breakatwhitespace=false,         
    breaklines=true,                 
    captionpos=b,                    
    keepspaces=true,                 
    numbers=left,                    
    numbersep=5pt,                  
    showspaces=false,                
    showstringspaces=false,
    showtabs=false,                  
    tabsize=2
}

\input{macros}

\title{A Library of LLM Intrinsics for \\ Retrieval-Augmented Generation}

\author{%
  Marina Danilevsky, Kristjan Greenewald, Chulaka Gunasekara, Maeda Hanafi, Lihong He,\\
  \textbf{Yannis Katsis, Krishnateja Killamsetty, Yulong Li, Yatin Nandwani, Lucian Popa, } \\
  \textbf{Dinesh Raghu, Frederick Reiss, Vraj Shah, Khoi-Nguyen Tran, Huaiyu Zhu, Luis Lastras} \\
  IBM Research\\
  \textit{June 2025}}

\date{April 2025}

\iclrfinalcopy

\begin{document}

\maketitle

 \begin{abstract}
   In the developer community for large language models (LLMs), there is not yet a clean pattern analogous to a software library, to support very large scale collaboration. Even for the commonplace use case of Retrieval-Augmented Generation (RAG), it is not currently possible to write a RAG application against a well-defined set of APIs that are agreed upon by different LLM providers. Inspired by the idea of compiler intrinsics, we propose some elements of such a concept through introducing a library of \textit{LLM Intrinsics} for RAG. An LLM intrinsic is defined as a capability that can be invoked through a well-defined API that is reasonably stable and independent of how the LLM intrinsic itself is implemented. The intrinsics in our library are released as LoRA adapters on HuggingFace, and through a software interface with clear structured input/output characteristics on top of vLLM as an inference platform, accompanied in both places with documentation and code. This article describes the intended usage, training details, and evaluations for each intrinsic, as well as compositions of multiple intrinsics.\footnote{This (June 2025) is the second version of this paper (the first was published in April 2025). Intrinsics implemented as LoRAs are now trained on IBM Granite 3.3 8b instruct (previously 3.2)}
 \end{abstract}

\section{Introduction}
\label{sec:introduction}

One of the most important software design patterns is the concept of a software library: generally reusable code with a well documented interface that enables very large scale collaboration between developers with different expertise. In large language models (LLMs), no such equivalent such pattern appears to have emerged as of yet. For example, prompt libraries tend to be useful only for a specific model. Even for the commonplace use case of Retrieval-Augmented Generation (RAG), it is not currently possible to write a RAG application against a well-defined set of APIs that are agreed upon by different LLM providers. Analogies to previous groundbreaking technologies abound; for example, different instruction set architectures used to be commonplace in microprocessor design, making code incompatible across such processors, and different operative systems offered different abstractions for applications that wanted to use system resources.


History suggests that the emergence of interfaces at key parts of  system design are inevitable, to allow different specializations to flourish and support the creation of more complex systems. 
The purpose of this article is to introduce the elements of a proposal in the context of RAG. We take inspiration from the idea of compiler intrinsics, which are functions that occur often enough to warrant inclusion in a programming language. The compiler is responsible for producing instructions that implement such functions in the specific computer architecture where software is expected to run, but it may take any leeway in optimizing such an implementation. 


In a loosely analogous concept, we define an LLM intrinsic to be a capability that can be invoked through a well-defined API that is reasonably stable and independent of how the LLM intrinsic itself is implemented. Metrics of performance, including accuracy, latency and throughput, may vary significantly across such implementations. We believe that LLM intrinsics are best implemented as a combination of a model and a co-optimized software layer that offers a familiar interface to the model developer. This pattern is already partly being followed in the LLM community; for example, LLM models in Huggingface are commonly packaged with configuration files for tokenizers, which transform structured representations of inputs (e.g., tool descriptions, sequences of messages) to raw tokens that are passed as actual inputs to the LLM. 



We present a library of RAG LLM intrinsics that are implemented both as LoRA adapters, and through a software interface with clear structured input/output characteristics on top of vLLM as an inference platform. For illustrative purposes, these intrinsics are implemented using IBM Granite language models, with extension to other model families possible in the future. We remark that nothing in the definition of an LLM intrinsic demands that it be built as an adapter; it could be implemented in a number of ways, including simply as part of the training data of the underlying model. This article is a sister article to \cite{greenewald2025alora}, which introduces the concept of activated LoRAs as a mechanism that can be used to implement LLM intrinsics in a highly inference-efficient way.


\subsection{Overview of the RAG LLM Intrinsics Library}
\label{sec:intrinsics_library}


The LLM Intrinsics RAG Library currently comprises eight intrinsics, each of which expects as input a (single-turn or multi-turn) conversation between a user and an AI assistant. Six of the intrinsics also expect a set of grounding passages. The functionality of each intrinsic is described below, and Table~\ref{tbl:intrinsics_hf} summarizes the inputs and outputs of each one.

\textbf{Query Rewrite (QR).} Given a conversation ending with a user query, QR will decontextualize that last user query by rewriting it (whenever necessary) into an equivalent version that is standalone and can be understood by itself. While this adapter is general purpose for any multi-turn conversation, it is especially effective in RAG settings where its ability to rewrite a user query into a standalone version directly improves the retriever performance, which in turn improves the answer generation performance.  This is a \textit{pre-retrieval} intrinsic since its suggested use is before invoking retrieval. 

\textbf{Query Expansion (QE).} Given a conversation ending with a user query, QE is designed to probe the retriever from multiple angles by generating a set of semantically diverse versions of that last user query. 
This expanded set of queries provides diverse retrieval paths, and thus this intrinsic is particularly effective in RAG settings, especially with terse, general, or underspecified queries. Like Query Rewrite, this is a \textit{pre-retrieval }intrinsic. 


\textbf{Context Relevance (CR).} Given a conversation ending with a user query, and an individual passage, CR classifies whether the passage is relevant, partially relevant, or irrelevant for answering the last user query - or if the passage may instead mislead or harm the downstream generator model's response quality. This is a \textit{pre-generation} intrinsic.

\textbf{Answerability Determination (AD).} Given a conversation ending with a user query, and a set of passages, AD classifies whether that final user query is answerable or unanswerable based on the available information in the passages. It is valuable for restraining over-eager models by identifying unanswerable queries and prevent the generation of hallucinated responses. It can also be used to indicate that the system should re-query the retriever with alternate formulations, to fetch more relevant passages. This is a \textit{pre-generation} intrinsic. 

\textbf{Passage Reranking (PRR).} Given a conversation ending with a user query, and a set of passages, PRR returns a ranked list of the passages ordered by suitability to answering the query. If the number of passages is small ($<10$) all passages are compared pairwise and returned ranked by win count; otherwise a tournament algorithm is used. This is a \textit{pre-generation} intrinsic.

\textbf{Uncertainty Quantification (UQ).} Given a conversation ending with an assistant response, UQ calculates a certainty percentage to reflect how certain it is about the  answer generated to the previous user query. UQ can also take as input a conversation ending with an user query and predicting the certainty score based solely on the query, prior to generating an answer. UQ is also calibrated on document-based question answering datasets, and hence it can be applied to giving certainty scores for RAG responses created using grounding passages. This intrinsic could be used in a \textit{post-generation} or \textit{pre-generation} step.

\textbf{Hallucination Detection (HD).} Given a conversation ending with an assistant response, and a set of passages, HD outputs a hallucination risk for each sentence in the last assistant response, with respect to the set of passages. It could be used in concert with sampling techniques that yield multiple generated responses, some of which could then be filtered according to their hallucination risks. This is a \textit{post-generation} intrinsic since its expected use is after invoking the LLM to create the response.

\textbf{Citation Generation (CG).} Given a conversation ending with an assistant response, and a set of passages, CG generates citations for that last assistant response from the provided passages. Citations are generated for each sentence in the response (when available), where each citation consists of a set of sentences from the supporting passages. This is a \textit{post-generation} intrinsic since its expected use is after invoking the LLM, and therefore can be used to create citations for responses generated by any model.

\begin{table}
    \small
    \centering
    \begin{tabular}{m{4.4cm}|
    >{\centering\arraybackslash}m{1.4cm}|
    >{\centering\arraybackslash}m{0.8cm}|
    >{\centering\arraybackslash}m{0.8cm}|
    >{\centering\arraybackslash}m{4cm}|
    >{\centering\arraybackslash}m{1cm}}
    \toprule
    
     & \multicolumn{3}{m{2cm}|}{\textbf{Input}} &  & \\
    \cmidrule{2-4}
    \textbf{Intrinsic} & \textbf{Passages} & \textbf{End Query} & \textbf{End Resp.} & \textbf{Output} & \textbf{Pre/Post}\\
    \midrule
    Query Rewrite (QR) & & $\times$ & & Standalone version of last query & Pre-R\\
    \midrule
    Query Expansion (QE) & & $\times$ & & Array of query variants & Pre-R\\
    \midrule
    Context Relevance (CR) & $\times$ (1~passage) & $\times$ & & Flag denoting if passage is relevant to query & Pre-G\\
    \midrule
    Answerability Determination (AD) & $\times$ & $\times$ & & Flag denoting if last query is answerable from passages & Pre-G\\
    \midrule
    Passage Reranking (PRR) & $\times$ & $\times$ & & Reranked list of passages & Pre-G\\
    \midrule
    \multirow{2}{4.4cm}{Uncertainty Quantification (UQ)} & $\times$  (optional) & $\times$ & & Certainty score for last assistant response (before generation) & Pre-G\\
    \cmidrule{2-6}
    & $\times$ (optional) &  & $\times$ & Certainty score for last assistant response (after generation) & Post-G\\
    \midrule
    Hallucination Detection (HD) & $\times$ &  & $\times$ & Hallucination risk for last assistant response & Post-G\\
    \midrule
    Citation Generation (CG) & $\times$ &  & $\times$ & Citations for last assistant response based on passages & Post-G\\
    \bottomrule
    \end{tabular}
  \caption{RAG LLM Intrinsics with their expected inputs and outputs. \emph{End Query} and \emph{End Resp.} refer to conversations ending with a user query and assistant response, respectively. \emph{Pre/Post} denotes if an intrinsic is called before retrieval (Pre-R), after retrieval and before generation (Pre-G), or after generation (Post-G).}
  \label{tbl:intrinsics_hf}
\end{table}

\subsection{RAG LLM Intrinsics Implementation}
\label{sec:intrinsics_library_release}

\renewcommand{\UrlFont}{\ttfamily\small}
Six of these intrinsics have been implemented by training a LoRA adapter for \url{ibm-granite/granite-3.3-8b-instruct} fine-tuned for a particular task. These have been released on HuggingFace as the Granite 3.3 8b RAG Agent Library, available at: \url{https://huggingface.co/ibm-granite/granite-3.3-8b-rag-agent-lib}. The six intrinsics thus implemented are Query Rewrite, Context Relevance, Answerability Determination, Uncertainty Quantification, Hallucination Detection, and Citation Generation.

However, the recommended use is via a second release mechanism: through \verb+Granite IO Processing,+\footnote{Granite IO can be found at: \url{https://github.com/ibm-granite/granite-io}} a framework which enables transforming how a user calls or infers an IBM Granite model and how the output from the model is returned to the user. In other words, the framework allows extended functionality of calling the model. This is particularly valuable as the downstream use of intrinsics relies on correctly structured output. Although we have made the individual LoRAs available, we strongly suggest that everyone uses the implementations in \verb+Granite IO+ and we have made example notebooks available.

Furthermore, the remaining two intrinsics, Query Expansion and Passage Reranking are pure software-based implementations and are only available through Granite IO. This emphasizes the general nature of intrinsics, where the implementation is abstracted from the function, as previously discussed.

In the rest of this paper we describe the specific implementation of each intrinsic in the library and evaluate their performance. We also discuss composing multiple intrinsics, and present particular implementations of composite flows accompanied by evaluations.

\input{query_rewrite}

\input{query_expansion}

\input{context_relevance}

\input{answerability}

\input{passage_reranking}

\input{uncertainty}

\input{hallucination}

\input{citations}

\section{Composite Intrinsics}
\label{sec:composite_intrinsics}

Individual intrinsics are created and trained to focus on particular tasks. In reality, we would certainly like to simultaneously improve retriever performance, reduce hallucinations, produce more accurate citations, and so on. Since the intrinsics' implementations are abstracted, it is simple to add one or more to a ``flow'' for a particular application.

For example, since using Query Rewrite improves recall performance, it is also likely to positively impact Citations, by providing more relevant contexts from which citations can be drawn. Or, intrinsics such as Uncertainty Quantification or Hallucination Detection could be combined with a sampling approach to response generation (such a sampling approach is incidentally available through \verb+Granite IO+) in order to easily filter out low quality candidates.

On the other hand, there are some composite flows that have a good chance of producing puzzling outcomes. For example, what might it mean if the same input yields a high score from Uncertainty Quantification (meaning, the model is quite certain about its answer) and yet low scores for  Hallucination Detection (meaning, the model believes the answer to be mostly unfaithful)? Or, what if a query is unanswerable according to Answerability Determination, and yet a subsequently generated answer is richly cited by Citation Generation? With every additional intrinsic added to an application flow, the complexity of testing and interpreting the resulting behavior significantly increases. Therefore, although many combinations may be technically possible, we recommend caution, and spend the rest of this section going through the process of creating and evaluating a composite intrinsic flow.

In particular, we will consider a flow which uses both the Query Rewrite (QR) and Answerability Determination (AD) intrinsics. These intrinsics are beneficial when the conversation with a RAG system is expected to frequently be multi-turn, and it is important to limit responses to only those which can be successfully supported (many customer-facing chat agents would fall under this use case). Although on the surface it may seem like neither of these intrinsics would affect each other's performance, we will see that the truth is a little more complicated.

\input{composite_qr_ad}

\section{Conclusion}
\label{conclusion}

In this paper we introduce a library of LLM intrinsics for RAG. Six intrinsics are currently implemented by training a LoRA adapter are Query Rewrite, Context Relevance, Answerability Determination, Uncertainty Quantification, Hallucination Detection, and Citation Generation.  These have been released on HuggingFace as the Granite 3.3 8b RAG Agent Library, available at: \url{https://huggingface.co/ibm-granite/granite-3.3-8b-rag-agent-lib}. Two additional intrinsics, Query Expansion and Passage Reranking, are pure software-based implementations and are only available through Granite IO: \url{https://github.com/ibm-granite/granite-io}. All the models are publicly released under an Apache 2.0 license for both research and commercial use. We describe the intended usage, training details, and evaluations for each intrinsic. We also introduce the notion of Composite Intrinsics, and describe one particular composition in detail, including in-depth evaluation of the created flow.

\section{Acknowledgments}
\label{acknowledgments}

Thanks to internal and external annotators.

\bibliographystyle{iclr2025_conference}
\bibliography{references}


\appendix

\section{Technical Appendices and Supplementary Material}

\input{appendices}

\end{document}

%% file: macros.tex
\newcommand{\quickstartpointer}{For an example describing how to use the intrinsic, refer to the documentation of the RAG agent library at \url{https://huggingface.co/ibm-granite/granite-3.3-8b-rag-agent-lib}.}

\newcommand{\quickstartpointergraniteio}{For an example describing how to use the intrinsic, refer to the relevant documentation in Granite IO at \url{https://github.com/ibm-granite/granite-io/tree/main/notebooks}.}

%% file: query_rewrite.tex
\section{Query Rewrite}
\label{sec:query_rewrite}

Granite 3.3 8b Instruct - Query Rewrite is a LoRA adapter for ibm-granite/granite-3.3-8b-instruct fine-tuned for the following task:

\begin{displayquote}
  Given a multi-turn conversation between a user and an AI assistant, decontextualize the last 
user utterance (query) by rewriting it (whenever necessary) into an equivalent version that 
is standalone and can be understood by itself.
\end{displayquote}

\subsection{Intended Use}

The query rewrite adapter is generally applicable for multi-turn conversational use cases. It is particularly useful in RAG settings where its ability to rewrite a user query into a standalone version directly improves the retriever performance, which in turn improves the answer generation performance. 

The rewrite is typically an expansion that in-lines, into the query, any implicit references that are made to entities, concepts, or even parts of the conversation that occur in the previous turns (either by the user or the AI assistant). Such expansion can include coreference resolution (i.e., replacement of pronouns with the actual entities), handling of ellipsis, which is the common linguistic phenomenon where parts of a sentence or phrase are omitted by the user, but can be understood from the context (i.e., for whom, of what, with respect to something discussed above, etc.).

As a result of the expansion, the query becomes a standalone query, still equivalent in meaning with what the user asked in the last turn. The rewritten query can be sent to downstream tasks (e.g., to a retriever in a RAG setting) as a better replacement for the original user query, and without the need for (a potentially very long) context.

\textbf{Input:} The input to the model is a list of conversational turns converted to a string using \verb+apply_chat_template+ function. These turns can alternate between the user and assistant roles, and the last turn is assumed to be from the user.

To prompt the LoRA adapter to rewrite the last user turn, a special rewrite role is used to trigger the rewrite capability of the model. The role includes the keyword "rewrite" followed by a short description of the query rewrite task.

\begin{lstlisting}
<|start_of_role|>rewrite: Reword the final utterance from the USER into a single utterance that doesn't need the prior conversation history to understand the user's intent. If the final utterance is a clear and standalone question, please DO NOT attempt to rewrite it, rather output the last utterance as is. Your output format should be in JSON: { \"rewritten_question\": <REWRITE> }"<|end_of_role|>
\end{lstlisting}

\textbf{Output:} When prompted with the above special rewrite role, the model generates a json object, which contains a field with the actual rewritten question.

Note: Even though one main application for query rewrite is in RAG settings, this LoRA adapter can be used to rewrite user questions for other conversational use cases (e.g., to access a database, or other APIs, or tools). As such, the adapter does not need any RAG documents (that may be present in the context, in a RAG setting) and uses only the dialog turns with what is being said between the user and assistant. 

\quickstartpointer

\subsection{Evaluation}

\subsubsection{Evaluation of the retriever}

We evaluate \verb+Recall@k+ on the MT-RAG benchmark \cite{katsis2025mtrag}, under various query rewrite strategies for the retriever. All retrieved passages are obtained using the Elser retriever with the same settings as in the above paper. In addition to the LoRA adapter, we include several other baselines, including no-rewrite (where we send the last user turn to the retriever as-is), Mixtral rewrites, as well as gold rewrites (human-created). We evaluate on three different testsets: a) full MT-RAG dataset (842 data points with last user turns); b) the non-standalone subset of MT-RAG dataset, which is a subset of 260 (out of 842) last user turns that were annotated by humans as non-standalone (i.e., they are dependent on the prior context); c) the standalone subset of MT-RAG dataset, which is the complementary subset, with all the last user turns that were annotated by humans as standalone.

Retrieval recall evaluation (Recall@k) with different query rewrite strategies, evaluated on full, non-standalone and standalone subsets of MT-RAG dataset are shown in Tables~\ref{table:rewrite_full}, \ref{table:rewrite_non_std}, and \ref{table:rewrite_std} respectively.
\begin{table}[h!]
\centering
\begin{tabular}{llll}
\toprule 
{\bf Rewrite Strategy}                             & {\bf Recall@5} & {\bf Recall@10} & {\bf Recall@20} \\
\midrule 
No rewrite                                   & 0.49     & 0.59      & 0.67      \\
Mixtral 8x7b                                 & 0.52     & 0.64      & 0.72      \\
Granite   3.3-8b-instruct-query-rewrite-LoRA & 0.56     & 0.68      & 0.76      \\
Gold rewrite                                 & 0.56     & 0.67      & 0.75     \\
\bottomrule 
\end{tabular}
\caption{Comparison of query rewrite strategies on the retrieval task of full MT-RAG dataset}
\label{table:rewrite_full}
\end{table}


\begin{table}[h!]
\centering
\begin{tabular}{llll}
\toprule 
{\bf Rewrite Strategy}                             & {\bf Recall@5} & {\bf Recall@10} & {\bf Recall@20} \\
\midrule 
No rewrite                                   & 0.26     & 0.39      & 0.44      \\
Mixtral 8x7b                                 & 0.36     & 0.49      & 0.57      \\
Granite   3.3-8b-instruct-query-rewrite-LoRA & 0.44     & 0.57      & 0.66     \\
Gold rewrite                                 & 0.48     & 0.58      & 0.66     \\
\bottomrule 
\end{tabular}
\caption{Comparison of query rewrite strategies on the retrieval task of non-standalone subset of MT-RAG}
\label{table:rewrite_non_std}
\end{table}

\begin{table}[h!]
\centering
\begin{tabular}{llll}
\toprule 
{\bf Rewrite Strategy}                             & {\bf Recall@5} & {\bf Recall@10} & {\bf Recall@20} \\
\midrule 
No rewrite                                   & 0.61     & 0.72      & 0.79      \\
Mixtral 8x7b                                 & 0.61     & 0.73      & 0.81      \\
Granite   3.3-8b-instruct-query-rewrite-LoRA & 0.63     & 0.75      & 0.83     \\
Gold rewrite                                 & 0.61     & 0.72      & 0.79     \\
\bottomrule 
\end{tabular}
\caption{Comparison of query rewrite strategies on the retrieval task of standalone subset of MT-RAG}
\label{table:rewrite_std}
\end{table}

If we focus on Recall@20 numbers, as one instance of the metric, there is an overall 9 percentage points jump when using query rewrite with the Granite 3.3-8b LoRA adapter versus when using the no rewrite strategy. This jump is more pronounced on the non-standalone fragment, where query rewrite with the Granite 3.3-8b LoRA adapter leads to 22 percentage points improvement over the no-rewrite strategy. Also, we can observe that the numbers with the LoRA rewrites are very close to what can be obtained with the gold rewrites on non-standalones (and slightly better on standalones for LoRA -- human annotators were instructed to leave the query unchanged when classifying it as standalone, however, the LoRA adapter may still perform some rewriting which turns out to further improve the recall).

\subsubsection{Evaluation of answer generation}

We evaluate answer generation quality, with top-k passages retrieved under the various query rewrite strategies for the retriever. We choose here k = 20, but similar trends take place for other values of k. We used Granite-3.3-8b instruct as the answer generator, and RAGAS Faithfulness (RAGAS-F) and RAD-Bench score as metrics for answer quality. We use the same three testsets as above.

The answer quality evaluation using RAGAS-F and RAD-Bench on full, non-standalone and standalone subsets of MT-RAG dataset are shown in Tables~\ref{table:rewrite_full_gen}, \ref{table:rewrite_non_std_gen}, and \ref{table:rewrite_std_gen} respectively.
\begin{table}[h!]
\centering
\begin{tabular}{llll}
\toprule 
{\bf Rewrite Strategy}                             & {\bf RAGAS-F} & {\bf RAD-Bench} \\
\midrule 
No rewrite                                   & 0.73     & 0.66      \\
Mixtral 8x7b                                 & 0.80     & 0.68      \\
Granite   3.3-8b-instruct-query-rewrite-LoRA & 0.81     & 0.70      \\
Gold rewrite                                 & 0.79     & 0.69     \\
\bottomrule 
\end{tabular}
\caption{Comparison of query rewrite strategies on the answer quality on full MT-RAG dataset}
\label{table:rewrite_full_gen}
\end{table}

\begin{table}[h!]
\centering
\begin{tabular}{llll}
\toprule 
{\bf Rewrite Strategy}                             & {\bf RAGAS-F} & {\bf RAD-Bench} \\
\midrule 
No rewrite                                   & 0.61     & 0.62      \\
Mixtral 8x7b                                 & 0.76     & 0.65      \\
Granite   3.2-8b-instruct-query-rewrite-LoRA & 0.79     & 0.69      \\
Gold rewrite                                 & 0.80     & 0.69     \\
\bottomrule 
\end{tabular}
\caption{Comparison of query rewrite strategies on the answer quality on non-standalone subset of MT-RAG}
\label{table:rewrite_non_std_gen}
\end{table}

\begin{table}[h!]
\centering
\begin{tabular}{llll}
\toprule 
{\bf Rewrite Strategy}                             & {\bf RAGAS-F} & {\bf RAD-Bench} \\
\midrule 
No rewrite                                   & 0.79     & 0.68      \\
Mixtral 8x7b                                 & 0.82     & 0.70     \\
Granite   3.3-8b-instruct-query-rewrite-LoRA & 0.83     & 0.71      \\
Gold rewrite                                 & 0.79     & 0.69     \\
\bottomrule 
\end{tabular}
\caption{Comparison of query rewrite strategies on the answer quality on standalone subset of MT-RAG}
\label{table:rewrite_std_gen}
\end{table}

As with Recall, similar observations can be made here as well. Specifically, we see an 8 percentage points jump in RAGAS Faithfulness and 4 percentage points jump in RAD-Bench score when using query rewrite with the Granite 3.3-8b LoRA adapter versus when using the no rewrite strategy. This improvement is more pronounced on the non-standalone fragment, where query rewrite with the Granite 3.3-8b LoRA adapter leads to a 18 percentage points jump in RAGAS Faithfulness and 7 percentage points jump in RAD-Bench score.

\subsection{Training Details}

The training data contains both: 1) standalone examples, which teach the adapter to refrain from rewriting user questions that are already standalone, and 2) non-standalone examples containing a diversity of patterns that are used to teach the adapter to expand the user turn so that it becomes standalone.

The training data uses the publicly available Cloud corpus of technical documentation pages from MT-RAG.\footnote{https://github.com/IBM/mt-rag-benchmark} Based on this corpus of documents, we constructed a dataset consisting of high-quality, human-created conversations, where the last turn of the conversation comes into versions: non-standalone version, and corresponding standalone version. The training dataset is proprietary and was obtained in combination with a third-party company who contracted the human annotators.

The LoRA adapter was fine-tuned using PEFT under the following regime: rank = $32$, learning rate = $3e-6$, number of epochs = $25$, with early stopping based on validation set, and $90/10$ split between training and validation.

%% file: query_expansion.tex
\section{Query Expansion}
\label{sec:query_expansion}

While Query Rewrite transforms a conversational query into a self-contained standalone form using a fine-tuned LoRA adapter, we introduce a Query Expansion intrinsic that takes this further by generating a set of semantically diverse queries designed to probe the retriever from multiple angles.  Instead of relying on a single rewrite, this intrinsic generates multiple candidate queries. These reflect different interpretations or formulations of the original user intent, improving the likelihood of retrieving relevant supporting passages. Given a multi-turn conversation, this intrinsic outputs a list of expanded queries by employing several expansion strategies: 
\begin{itemize}
    \item last user-turn question (no rewrites)
    \item Query Rewrite LoRA
    \item Answer Sampling: Samples a plausible answer using the granite instruct model without using any documents.
    \item Backward Generation from Answer: Reverse-engineer a potential question that could have elicited the answer.
    \item Synonymic Rewrite: Prompts Granite with explicit instructions to generate synonymous version of the query.
\end{itemize}

\subsection{INTENDED USE}

The query expansion intrinsic is designed for multi-turn conversational retrieval settings, particularly those involving RAG pipelines.  Its intended use is to generate multiple semantically varied formulations of the final user query, enabling more diverse retrieval. The expanded queries can be used independently to query a retriever.  The expanded queries provide diverse retrieval paths, improving the chances of surfacing relevant supporting content from the documents.

\textbf{Input:} The input to the model is a list of conversational turns. These turns can alternate between the user and assistant roles, and the last turn is assumed to be from the user.

\textbf{Output:} List of expanded queries (with user role).

\quickstartpointergraniteio

\subsection{EVALUATION}

\subsubsection{EVALUATION OF THE RETRIEVER}

We evaluate retrieval performance using Recall@k on the MT-RAG benchmark, comparing Query Expansion against a range of querying baselines in Table~\ref{table:query_expansion_retriever}. The retrieval backend is the Elser retriever, kept fixed across all experiments to isolate the impact of query formulation. With query expansion method, each expanded query is used to retrieve top-$20$ passages using Elser retriever. The final set of retrieved passages is obtained by taking set union across all querying methods, yielding a maximum of $100$ passages. 

\begin{table}[h!]
\centering
\begin{tabular}{lll}
\toprule 
{Method} & {Recall@50} & {Recall@100} \\
\midrule 
Last user-turn (No rewrite) & 74 & 77 \\
Query Rewrite LoRA & 84 & 87 \\
Answer-1 & 78 & 85 \\
Reverse-Engineered Question & 79 & 84 \\
Synonymous Query Rewrite & 62 & 69 \\
Query Expansion & 87 & 89 \\
\bottomrule  
\end{tabular}
\caption{Comparison of different querying strategies on the retrieval task on full MT-RAG dataset}
\label{table:query_expansion_retriever}
\end{table}

We observe that Query Expansion outperforms all single-query baselines, achieving Recall of 87\% with top-$50$ retrieved passages, compared to 84\% for the best-performing individual method (Query Rewrite LoRA). At Recall@100, the margin remains consistent, with Query Expansion reaching 89\%, showing that expanded queries retrieve relevant documents earlier. Beyond improvements in Recall@k, we observe distinct qualitative benefits of Query Expansion over traditional single-query methods: examples where the user's last-turn query is vague or underspecified. In such cases, a single rewrite often commits to one interpretation, which may miss relevant documents that align with alternate plausible meanings. By contrast, Query Expansion generates multiple semantically distinct versions of the query, each capturing a different potential reading of the user's intent.

\begin{table}[h!]
\centering
\begin{tabular}{lllll}
\toprule 
Method & RAGAS-F@50 & RAGAS-F@100 & RAD-Bench@50 & RAD-Bench@100 \\
\midrule
Last user-turn (No rewrite) & 77 & 82 & 66 & 66 \\
Query Rewrite LoRA & 83 & 86 & 68 & 67 \\
Answer-1 & 85 & 86 & 69 & 68 \\
Reverse-Engineered Question & 82 & 86 & 67 & 67 \\
Synonymous Query Rewrite & 75 & 80 & 67 & 66 \\
Query Expansion & 85 & 85 & 67 & 67\\
\bottomrule  
\end{tabular}
\caption{Comparison of different querying strategies on the answer quality on full MT-RAG dataset}
\label{table:query_expansion_generation}
\end{table}

\subsubsection{EVALUATION OF ANSWER GENERATION}

We evaluate answer generation quality, with top-k passages retrieved under the various querying strategies for the retriever. We used Granite3.3-8b instruct as the answer generator, and RAGAS Faithfulness (RAGAS-F) and RAD-Bench score as metrics for answer quality. Table~\ref{table:query_expansion_generation} shows the results. We find that methods like Answer-1 and Query Rewrite LoRA yield better faithfulness and correctness scores, indicating synergy in retrieval and generation scores. Last user-turn (No rewrite) shows the lowest generation quality scores related to other methods, highlighting the benefits of query modification strategies. Query Expansion maintains competitive performance. Importantly, it does not cause any regression in answer quality relative to other methods.

\textbf{Recommended usecases.}
\begin{itemize}
    \item This can be especially useful for ambiguous or underspecified queries, where a single rewrite may not fully capture user intent. 
    \item This is useful at first-turn of question where single query rewrite does nothing.
    \item The appropriate user would be paired with subset selection algorithm optimized for diversity, representativeness, etc.
\end{itemize}

%% file: context_relevance.tex
\section{Context Relevance}

Granite 3.3 8b Instruct - Context Relevance is a LoRA adapter for 
\verb+granite-3.3-8b-instruct+, that is fine-tuned for the context relevancy task: 

\begin{displayquote}
Given (1) a document and (2) a multi-turn conversation between a user and an AI assistant, identify whether the document is relevant (including partially relevant) and useful to answering the last user question.
\end{displayquote}

While this adapter is general-purpose and can even be used in cases where there is only one question, it is especially effective in RAG settings right after the retrieval model's step, where the adapter can be used to identify documents or passages that may mislead or harm the downstream generator model's response generation. 

\subsection{Intended Use}

This LoRA adapter enables classification of relevant, partially relevant, and irrelevant documents for the final user query in a multi-turn conversation. The model has been trained to determine whether a document is useful and relevant for answering the last user question in a multi-turn conversation.

The classification output from the context relevancy model can be used in several downstream applications, including but not limited to:

\begin{itemize}
    \item Filter out irrelevant and misleading documents before sending them to a generator model in a RAG setting. By removing irrelevant documents upfront, the downstream generator model reduces the likelihood of outputting misleading responses. Moreover, removing irrelevant documents or passages can help reduce the total downstream context length, thereby improving inference time.

    \item Signal to human annotators working in RAG settings which documents are irrelevant/relevant to the current turn of the conversation they are reviewing. Identifying such documents helps reduce the human annotator's high cognitive load (~\cite{hanafi2025rag}) involved in manually reading and reviewing several documents, especially in multi-turn conversations.

\end{itemize}

\textbf{Model Input:} The input to the model is a list of conversational turns and a list of documents, where each document is a dict containing the fields \verb+title+ and \verb+text+. The turns in the conversation can alternate between the user and the assistant, and the last turn is assumed to be from the user. For every document in the list of documents, the model converts that document and the conversation into a string using the \verb+apply_chat_template+ function.

To prompt the LoRA adapter to determine context relevancy, a special context relevancy role is used to trigger this capability of the model. The role includes the keyword ``context\_relevance'': \verb+<|start_of_role|>context_relevance<|end_of_role|>+

\textbf{Model Output:} When prompted with the above input, the model generates a JSON structure containing the context relevance output (``irrelevant'', ``partially relevant'', ``relevant''), as shown in the following example:

\begin{verbatim}
{
  "context_relevance": "irrelevant"
}
\end{verbatim}

\quickstartpointer

\subsection{Evaluation}

Our model was evaluated against several existing benchmarks to measure its ability to determine whether a document is relevant or not to the last user question in a multi-turn conversation.

\subsubsection{Forming evaluation datasets for context relevancy}

For each of the datasets, we created pairs of documents and questions and generated a label (based on the available labels in the dataset) as to whether the document is relevant or useful to answering the question. 

For cases where all documents and question pairs are positive labels (the document is useful/relevant for answering the question), an instance with a negative label is generated by randomly pairing the document with an irrelevant question.

Given the dataset ground truth and the LoRA outputs, we then calculated the precision and recall scores for the relevant and irrelevant data instances.

\subsubsection{Evaluation Datasets}

We used queries and document sets from the following datasets for our evaluation:

\begin{itemize}
    \item Multi-turn conversational RAG benchmarks: MTRAG (\cite{katsis2025mtrag}). The MT-RAG dataset comes with three settings of RAG retrieval: (1) Reference: perfect retriever (all passages are relevant to the question); (2) Reference + RAG: references passages + additional passages resulting in 5 total passages; (3) RAG: standard RAG setting (top 5 retrieved passages). We split the evaluation dataset according to the second and third settings and formed pairs of questions and passage sets. The dataset comes with pre-existing negative labels.

    \item Question Answering datasets, that involve \textit{implicit-fact retrieval}(\cite{zhao2024retrievalaugmentedgenerationrag}): CLAPNQ (\cite{rosenthal-etal-2025-clapnq}) and Drop (\cite{dua-etal-2019-drop}). The datasets contain pre-existing negative labels.

    \item Several RAG benchmarks that have previously been used to evaluate Granite models: FinanceBench (\cite{islam2023financebenchnewbenchmarkfinancial}), BioASQ\footnote{We use the subset from \url{https://huggingface.co/datasets/enelpol/rag-mini-bioasq}} (\cite{beir2021}), Open Australian Legal Corpus (\cite{butler-2023-open-australian-legal-dataset}). We generated negative labels by random pairings.

    \item Granite Guardian (\cite{padhi2024graniteguardian}) datasets for context relevancy: HotpotQA and SquadV2. The evaluation datasets contained pre-existing negative labels, which were obtained via prompting a separate model over the original datasets.

\end{itemize}

\subsubsection{Evaluation Results}

We evaluated the context relevance LoRA adapter against \verb+meta-llama/Llama-3.3-70B-Instruct+, \verb+ibm-granite/granite-guardian-3.1-5b+, and the Granite model \verb+ibm-granite/granite-3.3-8b-instruct+.  For Llama, we prompted the model using the prompt proven to perform best for context relevancy(~\cite{rahmani2025judgingjudgescollectionllmgenerated}).

\begin{table}[h!]
\centering
\begin{tabular}{p{4.5cm}|p{2cm}p{2cm}p{2cm}p{1.8cm}p{0.1cm}}
\toprule 
\multirow{2}{*}{\bf Evaluation Dataset} & {\bf Llama} & {\bf Granite} & {\bf Granite 3.3 } & {\bf Granite 3.3 } & \\
& {\bf 3.3 70B} & {\bf Guardian} & {\bf Instruct} & {\bf LoRA} &  \\
\midrule

    {\bf MTRAG} {\it(RAG)}  & 0.93 & 0.90 & 0.93 & \textbf{0.96} &  \\

    {\bf MTRAG} {\it(Reference + RAG)} & 0.87 & 0.83 & 0.86 &   \textbf{0.92} &  \\

    \hline

    {\bf CLAPNQ}  & 0.97 & 0.31 & 0.69 &   \textbf{0.99} &  \\

    \textbf{Drop} & 0.86 & 0.35 & 0.73 &  \textbf{0.94} & \\

    \hline

    {\bf FinanceBench} & 0.92 & 0.44 & 0.66 & \textbf{0.99} &   \\

     \textbf{BioASQ} & 0.98 & 0.83 & 0.94 &   \textbf{0.99} &  \\

     \textbf{Open Aus. Legal Corpus} & 0.93 & 0.08 & 0.60 &  \textbf{1.00} & \\

    \hline

    {\bf HotpotQA + SquadV2}  & 0.54 & 0.13 & 0.54 & \textbf{0.65} &  \\

\bottomrule 
\end{tabular}
\caption{Context relevance evaluation: Precision scores on the relevant labels}
\label{tbl:context-relevance-eval-precision}
\end{table}

\begin{table}[h!]
\centering
\begin{tabular}{p{4.5cm}|p{2cm}p{2cm}p{2cm}p{1.8cm}p{0.1cm}}
\toprule 
\multirow{2}{*}{\bf Evaluation Dataset} & {\bf Llama} & {\bf Granite} & {\bf Granite 3.3 } & {\bf Granite 3.3 } & \\
& {\bf 3.3 70B} & {\bf Guardian} & {\bf Instruct} & {\bf LoRA} &  \\
\midrule

    {\bf MTRAG} {\it(RAG)}  & 0.41 & 0.21 & 0.29 & \textbf{0.73} &  \\

    {\bf MTRAG} {\it(Reference + RAG)} & 0.39 & 0.21 & 0.29 & \textbf{0.73} &  \\

    \hline

    {\bf CLAPNQ}  &  0.97 & 0.00 & 0.65 & \textbf{0.99} &  \\

    \textbf{Drop} & 0.84 & 0.02 & 0.64 & \textbf{0.94} & \\

    \hline

    {\bf FinanceBench} & 0.89 & 0.00 & 0.34 & \textbf{0.99} &   \\

     \textbf{BioASQ} & 0.87 & 0.00 & 0.53 & \textbf{0.99} &  \\

     \textbf{Open Aus. Legal Corpus} & 0.92 & 0.00 & 0.36 &  \textbf{1.00} & \\

    \hline

    {\bf HotpotQA + SquadV2}  & 0.16 & 0.08 & 0.15 & \textbf{0.35} &  \\

\bottomrule 
\end{tabular}
\caption{Context relevance evaluation: Recall scores on the irrelevant labels}
\label{tbl:context-relevance-eval-recall}
\end{table}

We report the precision scores on the relevant labels (Table ~\ref{tbl:context-relevance-eval-precision}) and the recall scores on the irrelevant labels (Table ~\ref{tbl:context-relevance-eval-recall}). We observe that the LoRA's precision scores of the relevant labels improve on all the benchmarks compared to the precision scores on the Granite Instruct, Llama, and Granite Guardian models. We observe the same trend in the recall scores of the irrelevant labels for the LoRA adapter. Our evaluations show that the context relevance LoRA adapter is able to effectively identify relevant and irrelevant documents across several benchmarks.

\subsection{Training Details}

The training data was generated synthetically using a Mixtral model and documents from the CLAPNQ dataset (\cite{rosenthal-etal-2025-clapnq}). Based on this corpus, we generated several pairs of documents and multi-turn conversations along with a relevant/partially relevant/irrelevant label indicating the document's relevance to the last user question. We then used Mixtral as an automatic judge to validate the generated labels and filter out noisy samples.

The LoRA adapter was fine-tuned using PEFT under the following regime: rank = 32, learning rate = 5e-6, number of epochs = 25.

%% file: answerability.tex
\section{Answerability Determination}
\label{sec:answerability}

Granite 3.3 8b Instruct - Answerability Determination is a LoRA adapter for ibm-granite/granite-3.3-8b-instruct fine-tuned for binary answerability classification task. The model takes as input a multi-turn conversation and a set of documents, and classifies whether the user's final query is answerable or unanswerable based on the available information in the set of input documents.

\subsection{Intended Use}

This is a LoRA adapter that enables answerability classification for the final user query in a multi-turn conversation, with respect to a set of provided documents. The model is trained to determine whether the last user query is answerable or unanswerable, based solely on the information present in the input documents. This makes it suitable for applications involving RAG and document-grounded chatbots, where knowing whether sufficient information exists to answer a query is crucial. The classification output from the answerability model can be used in several downstream applications, including but not limited to:

\begin{itemize}
    \item     Filter out unanswerable questions before sending them to generation in RAG setting. By classifying a query as unanswerable upfront, the system can prevent hallucinated or misleading responses.
    \item     Re-query the retriever to get more relevant documents. If a query is initially deemed unanswerable, the retriever can be re-invoked with alternate formulations to fetch more relevant documents.
\end{itemize}

\textbf{Model input:} The input to the model is a list of conversational turns and a list of documents converted to a string using \verb+apply_chat_template+ function. These turns can alternate between the user and assistant roles. The last turn is from the user. The list of documents is a dictionary with text field, which contains the text of the corresponding document.

To prompt the LoRA adapter to determine answerability, a special answerability role is used to trigger this capability of the model. The role includes the keyword \verb+"answerability": <|start_of_role|>answerability<|end_of_role|>+

\textbf{Model output:} When prompted with the above input, the model generates the answerable or unanswerable output.

\quickstartpointer

\subsection{Evaluation}

\subsubsection{Answerability Classification }

We evaluated the model against baselines on binary answerability classification using two separate benchmarks:
\begin{itemize}
    \item
    Single-turn Setting (SQUADRun Benchmark~\cite{rajpurkar2018know}): In this setting, the user query and the supporting documents are provided. Our model was evaluated against standard baselines to measure its ability to determine whether a standalone question is answerable based on the document set. Table~\ref{table:ans_class_squad} shows the classification results.



\begin{table}[t]
\centering
\begin{tabular}{l|c|c|c|c|c|c|c}
\toprule 
\textbf{Model} & \makecell{\textbf{Unans.} \\ \textbf{Precision}} & \makecell{\textbf{Unans.} \\ \textbf{Recall}} & \makecell{\textbf{Unans.} \\ \textbf{F1}} & \makecell{\textbf{Ans.} \\ \textbf{Precision}} & \makecell{\textbf{Ans.} \\ \textbf{Recall}} & \makecell{\textbf{Ans.} \\ \textbf{F1}}  & \makecell{\textbf{Weighted} \\ \textbf{F1}} \\
\midrule 
BigBird w/ MLP & 49.2 & 68.5 & 57.3 & 48.0 & 29.2 & 36.3 & 46.8 \\
LLaMA 2-7B & 72.2 & 71.0 & 71.6 & 71.4 & 72.6 & 72.0 & 71.8 \\
Granite 3.3-8b LoRA & 88.1 & 59.3 & 70.9 & 69.3 & 92 & 79 & 75 \\
\bottomrule 
\end{tabular}
\caption{Comparison of classification performance across models on SQUADRUN Dev set. Metrics are broken down by class (Answerable vs. Unanswerable) and include precision, recall, and F1 score.}
\label{table:ans_class_squad}
\end{table}

    \item
    Multi-turn Setting (MT-RAG Benchmark~\cite{katsis2025mtrag}): In this setting, the model is given the full multi-turn conversation history along with the supporting documents. This benchmark evaluates the model's ability to assess answerability when the final user query can also depend on prior turns for context. Table~\ref{table:ans_class_mtrag} shows the results.
\end{itemize}
    

\begin{table}[t]
\centering
\begin{tabular}{l|c|c|c|c|c|c|c}
\toprule 
\textbf{Model} & \makecell{\textbf{Unans.} \\ \textbf{Precision}} & \makecell{\textbf{Unans.} \\ \textbf{Recall}} & \makecell{\textbf{Unans.} \\ \textbf{F1}} & \makecell{\textbf{Ans.} \\ \textbf{Precision}} & \makecell{\textbf{Ans.} \\ \textbf{Recall}} & \makecell{\textbf{Ans.} \\ \textbf{F1}}  & \makecell{\textbf{Weighted} \\ \textbf{F1}} \\
\midrule 
BigBird w/ MLP & 69.6 & 77.6 & 73.4 & 70.1 & 60.8 & 65.2 & 69.6 \\
LLaMA 2-7B & 86.9 & 89.4 & 88.2 & 87.3 & 84.5 & 85.9 & 87.1 \\
Granite 3.3-8b LoRA & 89.8 & 91.8 & 90.8 & 90.3 & 87.9 & 89.1 & 90 \\
\bottomrule 
\end{tabular}
\caption{Comparison of classification performance across models on MT-RAG Benchmark. Metrics are broken down by class (Answerable vs. Unanswerable) and include precision, recall, and F1 score.}
\label{table:ans_class_mtrag}
\end{table}

\subsubsection{Comparing LoRA Adapter vs. Vanilla Granite for Answer Quality}

We compare the performance of Granite 3.3-8b Instruct vs. Granite 3.3-8b LoRA adapter on a subset of MT-RAG Benchmark in Table~\ref{table:ans_class_generation}. In this setup, each query is paired with only 5 retrieved passages as context. The true answerability label for each query indicates whether the query is answerable with respect to the retrieved context.

    \begin{itemize}

    \item
    Answerability Classification Performance: The LoRA adapter outperforms the vanilla model in overall F1 on both answerables and unanswerables. The LoRA adapter achieves higher recall on unanswerable queries, making it better at identifying questions that should not be answered. However, this comes at the cost of lower recall on answerable queries.


    \item
    Joint Answerability-Faithfulness Score (\texttt{JAFS)}:
    \[
    \texttt{JAFS} =
    \begin{cases}
    1 & \text{if prediction = IDK/unanswerable \& truth = unanswerable} \\
    \texttt{RF} & \text{if prediction = non-IDK/answerable \& truth = answerable} \\
    0 & \text{otherwise}
    \end{cases}
    \]

    This score rewards the model for correctly abstaining on unanswerable queries (full credit) and for providing faithful answers on answerable queries (partial credit based on RAGAS Faithfulness). No credit is given for incorrect or unfaithful predictions.

The LoRA adapter achieves a 17\% lift on this metric - rewarding the model for correctly abstaining on unanswerable queries and for being faithful when it chooses to answer.

\end{itemize}


\begin{table}[h]
\centering
\begin{tabular}{l|c|c|c|c|c|c}
\toprule 
\textbf{Model} & \makecell{\textbf{Unans.} \\ \textbf{F1}} & \makecell{\textbf{Ans.} \\ \textbf{F1}}  & \makecell{\textbf{Unans.} \\ \textbf{Recall}} & \makecell{\textbf{Ans.} \\ \textbf{Recall}} &
\makecell{\textbf{JAFS}} \\
\midrule
\textbf{Granite 3.3-8b Instruct} & 17 & 77 & 10 & 99 & 49 \\
\textbf{Granite 3.3-8b LoRA} & 69 & 81 & 67 & 82 & 66 \\
\bottomrule
\end{tabular}
\caption{Comparison of Granite 3.3-8B Instruct vs. LoRA Adapter on Answerability and Faithfulness metrics using MT-RAG Benchmark.}
\label{table:ans_class_generation}
\end{table}

\subsection{Training Details}

The training data uses the publicly available Government corpus from MT-RAG\cite{katsis2025mtrag} as the source of documents. Based on this corpus, we constructed a dataset consisting of a mix of human-created and synthetically generated multi-turn conversations. It includes two types of examples: (1) Answerable queries, where the final user question can be answered based on the provided documents. These examples teach the adapter to recognize when sufficient information is present to support an answer. (2) Unanswerable queries, where the documents lack the necessary information to answer the final user query. We used Mixtral as an automatic judge to validate the answerability labels and filter out noisy samples.

The LoRA adapter was fine-tuned using PEFT under the following regime: rank = 32, learning rate = 5e-6, number of epochs = 25, with early stopping based on validation set, and 90/10 split between training and validation.

%% file: passage_reranking.tex
\section{Passage Reranking}
\label{sec:passage_reranking}

Granite 3.3 Instruct - Passage Reranking is a prompt-based intrinsic for reranking retrieved passages.  It takes the output of the retrieval step as input and returns a reranked (subset of the) retrieved passages which can be then used for generation.

\subsection{INTENDED USE}

The reranking is done in a tournament style by prompting an LLM to compare paired passages and return the label of the preferred one. Users can provide their own prompt; otherwise the following default prompt is used:

\begin{lstlisting}
You are a smart and helpful AI assistant with in-depth knowledge about how people search for information using search engines. In this task, you are  given two passages, and a query, your job is to judge which passage is relatively more suitable to answer that query. The first passage will start with "passage A" and the  second passage with start with "passage B". Output the preferred passage index, i.e. A  or B and followed by an explanation, if none of the passage answer the query directly,  pick the one that has more relevant information.
\end{lstlisting}

When the number of reranking passages is less than 10, the intrinsic compares all possible pairs and ranks by each passage's win count. All or multiple pairs are grouped into single batch for efficiency. When the number of reranking passages is larger than 10, the intrinsic forms pairs using the original rankings and only advances the winning passage from each pair to the next round. If there are an odd number of documents, the last document is dropped for simplicity. 

\quickstartpointergraniteio

\subsection{EVALUATION}

We provide the re-ranking result for the popular BEIR (\cite{beir2021}) benchmark as reference. By closely examining the re-ranking results, we find that many of the re-ranked top passages across different tasks are at least as good as the gold passages, although they are not counted as gold. We suspect this contributes to the performance drop in some tasks.

\begin{table}[ht]
\centering
\begin{tabular}{lcc}
\toprule
\textbf{NDCG@10}      & \textbf{granite-embedding-30M} & \textbf{Granite 3.3 Instruct (rerank top 40)} \\
\midrule
trec-covid & 	0.631 & 	0.724 \\ 
nq & 	0.516 & 	0.586 \\ 
fiqa & 	0.369 & 	0.371 \\ 
scifact & 	0.713 & 	0.742 \\ 
msmarco & 	0.307 & 	0.322 \\ 
hotpotqa & 	0.629 & 	0.652 \\ 
nfcorpus & 	0.337 & 	0.367 \\ 
touche2020 & 	0.240 & 	0.223 \\ 
scidoc & 	0.225 & 	0.210 \\ 
dbpedia & 	0.360 & 	0.405 \\ 
fever & 	0.855 & 	0.848 \\ 
climatefever & 	0.303 & 	0.301 \\ 
CQADupstaceRetrieval & 	0.443 & 	0.459 \\ 
quora & 	0.867 & 	0.805 \\ 
arguana & 	0.567 & 	0.347 \\
\bottomrule
\end{tabular}
\caption{NDCG@10 scores on the retrieval tasks in the BEIR benchmark, prompting Granite 3.3 Instruct.}
\label{tab:mae_pre_post}
\end{table}

Unlike most generation tasks, reranking only needs to ouput 1 single preferance token, A or B. Therefore, the processing time is significantly less than regular generation and it only adds a small overhead in the end to end RAG pipeline. Possible future work includes support for reranking a set of passages rather than only a pair.

%% file: uncertainty.tex
\section{Uncertainty Quantification}
\label{sec:uncertainty}

Granite 3.3 8b Instruct - Uncertainty Quantification is a LoRA adapter for ibm-granite/granite-3.3-8b-instruct, adding the capability to provide calibrated certainty scores when answering questions when prompted, in addition to retaining the full abilities of the ibm-granite/granite-3.3-8b-instruct model.
The model is a LoRA adapter finetuned to provide certainty scores mimicking the output of a calibrator trained via the method in \cite{shen2024thermometeruniversalcalibrationlarge}.

\subsection{Intended Use}

\textbf{Certainty score definition.} The model will respond with a certainty percentage, quantized to 10 possible values (i.e. 5\%, 15\%, 25\%,...95\%). This percentage is calibrated in the following sense: given a set of answers assigned a certainty score of X\%, approximately X\% of these answers should be correct. See the eval experiment below for out-of-distribution verification of this behavior.

\textbf{Certainty score interpretation.} Certainty scores calibrated as defined above may at times seem biased towards moderate certainty scores for the following reasons. Firstly, as humans we tend to be overconfident in our evaluation of what we know and don't know - in contrast, a calibrated model is less likely to output very high or very low confidence scores, as these imply certainty of correctness or incorrectness. Examples where you might see very low confidence scores might be on answers where the model's response was something to the effect of "I don't know", which is easy to evaluate as not being the correct answer to the question (though it is the appropriate one). Secondly, remember that the model is evaluating itself - correctness/incorrectness that may be obvious to us or to larger models may be less obvious to an 8b model. Finally, teaching a model every fact it knows and doesn't know is not possible, hence it must generalize to questions of wildly varying difficulty (some of which may be trick questions!) and to settings where it has not had its outputs judged. Intuitively, it does this by extrapolating based on related questions it has been evaluated on in training - this is an inherently inexact process and leads to some hedging.

Important note: Certainty is inherently an intrinsic property of a model and its abilities. Granite-3.3-8b-Uncertainty-Quantification is not intended to predict the certainty of responses generated by any other models besides itself or ibm-granite/granite-3.3-8b-instruct. Additionally, certainty scores are distributional quantities, and so will do well on realistic questions in aggregate, but in principle may have surprising scores on individual red-teamed examples.

\subsubsection{Usage steps}

Note that the model was trained with 1/3 each of the data having (a) no system prompt, (b) the default generic Granite system prompt, and (c) the default Granite RAG prompt. As a result, the model is robust to these choices of system prompt.

There are two supported usage scenarios.

\textbf{Scenario 1.} Answering a question and obtaining a certainty score proceeds as follows. Given a user query written in the user role:

\begin{enumerate}
    \item     Use the base model to generate a response as normal (via the assistant role).
    \item     Prompt the model to generate a certainty score by generating in the certainty role (use "certainty" as the role in the chat template, or simply append \verb+<|start_of_role|>certainty<|end_of_role|>+ and continue generating).
    \item     The model will respond with a certainty percentage, quantized with steps of 10\% (i.e. 05\%, 15\%, 25\%,...95\%). Note, any additional text after the score and \% can be ignored. You can curb additional generation by setting "max token length" = 3 when using this role.
\end{enumerate}

\textbf{Scenario 2.} Predicting the certainty score from the question (optionally plus documents) only, prior to generating an answer. Given a user query written in the user role:

\begin{enumerate}
    \item     Prompt the model to generate a certainty score by generating in the certainty role (use "certainty" as the role in the chat template, or simply append \verb+<|start_of_role|>certainty<|end_of_role|>+ and continue generating).
    \item     The model will respond with a certainty percentage, quantized with steps of 10\% (i.e. 05\%, 15\%, 25\%,...95\%). Note, any additional text after the score and \% can be ignored. You can curb additional generation by setting "max token length" = 3 when using this role.
    \item     Remove the generated certainty string, and if desired, use the base model to generate a response as normal (via the assistant role).
\end{enumerate}

\quickstartpointer

\subsubsection{Possible downstream use cases (not implemented)}

\begin{itemize}
    \item     Human usage: Certainty scores give human users an indication of when to trust answers from the model (which should be augmented by their own knowledge).
    \item     Model routing/guards: If the model has low certainty (below a chosen threshold), it may be worth sending the request to a larger, more capable model or simply choosing not to show the response to the user.
    \item     RAG: Granite-3.3-8b-Uncertainty-Quantification is calibrated on document-based question answering datasets, hence it can be applied to giving certainty scores for answers created using RAG. This certainty will be a prediction of overall correctness based on both the documents given and the model's own knowledge (e.g. if the model is correct but the answer is not in the documents, the certainty can still be high).
\end{itemize}    

\subsection{Evaluation}

The model was evaluated on the MMLU\footnote{https://huggingface.co/datasets/cais/mmlu} datasets (not used in training). Shown are the Expected Calibration Error (ECE)\footnote{https://towardsdatascience.com/expected-calibration-error-ece-a-step-by-step-visual-explanation-with-python-code-c3e9aa12937d} for each task, for the base model (Granite-3.3-8b-instruct) and Granite-3.3-8b-Uncertainty-Quantification. The average ECE across tasks for our method is 0.064 (out of 1) and is consistently low across tasks (maximum task ECE 0.10), compared to the base model average ECE of 0.20 and maximum task ECE of 0.60. Note that our ECE of 0.064 is smaller than the gap between the quantized certainty outputs (10\% quantization steps). Additionally, the zero-shot performance on the MMLU tasks does not degrade, averaging at 89\%.

\begin{figure}
    \centering
    \includegraphics[width=0.5\linewidth]{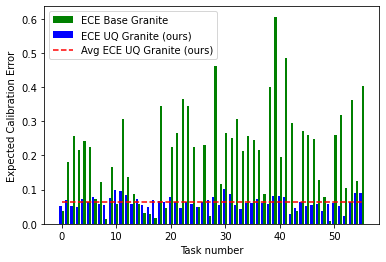}
    \caption{Evaluation of UQ Intrinsic}
    \label{fig:uncertainty_eval}
\end{figure}

\paragraph{System prompt and application point} Table \ref{tab:mae_pre_post} shows results of predicting the target certainty score before (Scenario 2) and after (Scenario 1) the assistant response under different prompting conditions. Note that the performance is robust to choice of system prompt (as all 3 are included in training). As should be expected, the Pre-answer (Scenario 2) performance is degraded, since it is predicting the certainty of the assistant answer before seeing the answer. That said, the MAE is still less than 1, i.e. less than the gap between adjacent certainty scores (integers 0 to 9).

\begin{table}[ht]
\centering
\begin{tabular}{lcc}
\toprule
\textbf{Prompt Condition}      & \textbf{MAE Post Answer} & \textbf{MAE Pre Answer} \\
\midrule
No system prompt               & 0.375                    & 0.915                   \\
Standard system prompt         & 0.360                    & 0.840                   \\
RAG prompt                     & 0.385                    & 0.840                   \\
\bottomrule
\end{tabular}
\caption{Mean Absolute Error (MAE) of predicting the target certainty score before (Scenario 2) and after (Scenario 1) the assistant response under different prompting conditions.  }
\label{tab:mae_pre_post}
\end{table}

\subsection{Training Details}

The model is a LoRA adapter finetuned to provide certainty scores mimicking the output of a calibrator trained via the method in \cite{shen2024thermometeruniversalcalibrationlarge}.

The following datasets were used for calibration and/or finetuning:

\renewcommand{\UrlFont}{\ttfamily\tiny}

\begin{itemize}
    \item BigBench (\url{https://huggingface.co/datasets/tasksource/bigbench})
    \item MRQA (\url{https://huggingface.co/datasets/mrqa-workshop/mrqa})
    \item newsqa (\url{https://huggingface.co/datasets/lucadiliello/newsqa})
    \item trivia\_{qa} (\url{https://huggingface.co/datasets/mandarjoshi/trivia_qa})
    \item search\_{qa} (\url{https://huggingface.co/datasets/lucadiliello/searchqa})
    \item openbookqa (\url{https://huggingface.co/datasets/allenai/openbookqa})
    \item web\_{questions} (\url{https://huggingface.co/datasets/Stanford/web_questions})
    \item smiles-{qa} (\url{https://huggingface.co/datasets/alxfgh/ChEMBL_Drug_Instruction_Tuning})
    \item orca-{math} (\url{https://huggingface.co/datasets/microsoft/orca-math-word-problems-200k})
    \item ARC-Easy (\url{https://huggingface.co/datasets/allenai/ai2_arc})
    \item commonsense\_qa (\url{https://huggingface.co/datasets/tau/commonsense_qa})
    \item social\_iq\_a (\url{https://huggingface.co/datasets/allenai/social_i_qa})
    \item super\_glue (\url{https://huggingface.co/datasets/aps/super_glue})
    \item figqa (\url{https://huggingface.co/datasets/nightingal3/fig-qa})
    \item riddle\_sense (\url{https://huggingface.co/datasets/INK-USC/riddle_sense})
    \item ag\_news (\url{https://huggingface.co/datasets/fancyzhx/ag_news})
    \item medmcqa (\url{https://huggingface.co/datasets/openlifescienceai/medmcqa})
    \item dream (\url{https://huggingface.co/datasets/dataset-org/dream})
    \item codah (\url{https://huggingface.co/datasets/jaredfern/codah})
    \item piqa (\url{https://huggingface.co/datasets/ybisk/piqa})
\end{itemize}

\renewcommand{\UrlFont}{\ttfamily\normalsize}

%% file: hallucination.tex
\section{Hallucination Detection}
\label{sec:hallucination}

Granite 3.3 8b Instruct - Hallucination Detection is a LoRA adapter for ibm-granite/granite-3.3-8b-instruct fine-tuned for the hallucination detection task of model outputs. Given a multi-turn conversation between a user and an AI assistant ending with an assistant response and a set of documents/passages on which the last assistant response is supposed to be based, the adapter outputs a hallucination risk for each sentence in the assistant response.

\subsection{Intended Use}

This is a LoRA adapter that gives the ability to identify hallucination risks for the sentences in the last assistant response in a multi-turn RAG conversation based on a set of provided documents/passages.

While you can invoke the LoRA adapter directly, we highly recommend calling it through \verb+Granite IO+, as described in Section \ref{sec:intrinsics_library_release}. \verb+Granite IO+ wraps the hallucination detection adapter with a tailored I/O processor. The I/O processor provides a friendlier development interface, as it takes care of various data transformations and validation tasks. This includes splitting the assistant response into sentences before calling the adapter, as well as validating the adapter's output and transforming the sentence IDs returned by the adapter into appropriate spans over the response. 

However, if you prefer to invoke the LoRA adapter directly, its expected input/output is described below.

\textbf{Model input:} The input to the model is conceptually a list of conversational turns ending with an assistant response and a list documents converted to a string using \verb+apply_chat_template+ function. For the adapter to work, the last assistant response should be pre-split into sentences and sentence indices need to be prepended. In more detail, the primary inputs are the following three items, each represented in JSON:

\begin{itemize}
    \item     \textbf{conversation}: A list of conversational turns between the user and the assistant, where each item in the list is a dictionary with fields \verb+role+ and \verb+content+. The \verb+role+ equals to either \verb+user+ or \verb+assistant+, denoting user and assistant turns, respectively, while the \verb+content+ field contains the corresponding user/assistant utterance. The conversation should end with an assistant turn and the \verb+text+ field of that turn should contain the assistant utterance with each sentence prefixed with a response sentence id of the form \verb+<iI>+, where \verb+I+ is an integer. The numbering should start from 0 (for the first sentence) and be incremented by one for each subsequent sentence in the last assistant turn.
    \item     \textbf{instruction}: A task instruction, which is encoded as a dictionary with fields \verb+role+ and \verb+content+, where \verb+role+ equals to \verb+system+ and \verb+content+ equals to the following string describing the hallucination detection task: "Split the last assistant response into individual sentences. For each sentence in the last assistant response, identify the faithfulness by comparing with the provided documents and generate the faithfulness reasoning and faithfulness decision. Ensure that your output includes all response sentence IDs, and for each response sentence ID, provide the corresponding faithfulness reasoning and faithfulness decision. The output must be a json structure."
    \item     \textbf{documents}: A list of documents, where each item in the list is a dictionary with fields \verb+doc_id+ and \verb+text+. The \verb+text+ field contains the text of the corresponding document.
\end{itemize}

To prompt the LoRA adapter, we combine the above components as follows: We first append the \verb+instruction+ to the end of the \verb+conversation+ to generate an \verb+augmented_conversation+ list. Then we invoke the \verb+apply_chat_template+ function with parameters: \verb+conversation = augmented_conversation+ and \verb+documents = documents+.

\textbf{Model output:} When prompted with the above input, the model generates a json structure which contains a list of dictionaries, where the dictionaries in the list correspond to the sentences in the last assistant output. A dictionary contains 3 fields, 

\begin{enumerate}
   \item \texttt{i} - sentence id of the last assistant turn.
   \item \texttt{f} - faithfulness/hallucination label for the sentence. This can take 4 values:
   \begin{enumerate}
     \item \texttt{faithful} - the sentence is faithful to the provided documents.
     \item \texttt{unfaithful} - the sentence is not faithful to the provided documents.
     \item \texttt{partial} - the sentence contain some claims that are not grounded to the contents of the documents.
     \item \texttt{NA} - the sentence does not contain any claims.
     \end{enumerate}
    \item \texttt{r} - reasoning for the the model's faithfulness label.
\end{enumerate}

\quickstartpointer

\subsection{Evaluation}
The LoRA adapter was evaluated on the QA portion of the RAGTruth benchmark \cite{niu2024ragtruth}. We compare the response-level hallucination detection performance between the LoRA adapter and the methods reported in the RAGTruth paper. The responses that obtain a faithfulness score less than $0.1$ for at least one sentence are considered as hallucinated responses.

The evaluation results are shown in the Table~\ref{table:hallucination_eval}. The results for the baselines are extracted from the RAGTruth paper \cite{niu2024ragtruth}.
\begin{table}[h!]
\centering
\begin{tabular}{llll}
\toprule 
{\bf Model}           & {\bf Precision} & {\bf Recall} & {\bf F1} \\
\midrule 
gpt-3.5-turbo (prompted)                                             & 18.8               & 84.4            & 30.8        \\
gpt-4-turbo (prompted)                                               & 33.2               & 90.6            & 45.6        \\
SelfCheckGPT \cite{manakul2023selfcheckgpt} & 35.0                & 58              & 43.7        \\
LMvLM \cite{cohen2023lm}        & 18.7               & 76.9            & 30.1        \\
Finetuned Llama-2-13B                                                & 61.6               & 76.3            & 68.2        \\
Hallucination Detection LoRA                                         & 68.1              & 69.2            & 68.6         \\
\bottomrule 
\end{tabular}
\caption{Hallucination detection results}
\label{table:hallucination_eval}
\end{table}

\subsection{Training Details}

The process of generating the training data consisted of two main steps:

\begin{itemize}
\item \textbf{Multi-turn RAG conversation generation:} Starting from publicly available document corpora, we generated a set of multi-turn RAG data, consisting of multi-turn conversations grounded on passages retrieved from the corpora. For details on the RAG conversation generation process please refer to the Granite Technical Report\footnote{https://github.com/ibm-granite/granite-3.0-language-models/blob/main/paper.pdf} as well as \cite{lee2024multidocumentgroundedmultiturnsynthetic}.
\item \textbf{Faithfulness label generation:} For each turn of the multi-turn RAG conversations from the previous step, we used a multi-step synthetic faithfulness label and reasoning generation pipeline to generate faithfulness labels for the assistant response.

\end{itemize}

The following public datasets were used as seed datasets for the multi-turn RAG conversation generation process:

\renewcommand{\UrlFont}{\ttfamily\tiny}

\begin{itemize}
\item CoQA Wikipedia Passages (\url{https://stanfordnlp.github.io/coqa/})
\item MultiDoc2Dial (\url{https://huggingface.co/datasets/IBM/multidoc2dial})
\item QuAC (\url{https://huggingface.co/datasets/allenai/quac})
\end{itemize}

\renewcommand{\UrlFont}{\ttfamily\normalsize}

The LoRA adapter was fine-tuned using PEFT under the following regime: rank = 16, learning rate = 1e-5, and 90/10 split between training and validation.

%% file: citations.tex
\section{Citation Generation}
\label{sec:citations}

Granite 3.3 8b Instruct - Citation Generation is a RAG-specific LoRA adapter for ibm-granite/granite-3.3-8b-instruct fine-tuned for the citation generation task. Given a multi-turn conversation between a user and an AI assistant ending with an assistant response and a set of documents/passages on which the last assistant response is supposed to be based, the adapter generates citations for the last assistant response from the provided documents/passages. The LoRA adapter has the following features:

\begin{itemize}
\item \textbf{Fine-grained citations:} The adapter generates citations for each sentence in the assistant response (when available). Moreover, each citation consists of a set of sentences from the documents/passages that support the corresponding sentence in the assistant response.   
\item \textbf{Post-hoc citation generation:} Since the adapter takes the assistant response as input, it can generate citations for responses generated by any LLM. Pick your favorite LLM and use the adapter to generate post-hoc citations!
\end{itemize}

\subsection{Intended Use}

This is a LoRA adapter that gives the ability to generate citations for the last assistant response in a multi-turn RAG conversation based on a set of provided documents/passages. It can be used to generate post-hoc citations for assistant responses generated by any LLM in a RAG setting.  

While you can invoke the LoRA adapter directly, we highly recommend calling it through \verb+Granite IO+, as described in Section \ref{sec:intrinsics_library_release}. \verb+Granite IO+ wraps the adapter with a tailored I/O processor. The I/O processor provides a friendlier development interface, as it takes care of various data transformations and validation tasks. This includes, among others, splitting the input documents and assistant response into sentences before calling the adapter, as well as validating the adapter's output and transforming the sentence IDs returned by the adapter into appropriate spans over the documents and the response. 

However, if you prefer to invoke the LoRA adapter directly, the expected input/output is described below.

\textbf{Model input:} The input to the model is conceptually a list of conversational turns ending with an assistant response and a list of documents converted to a string using the \verb+apply_chat_template+ function. For the adapter to work, the last assistant response as well as the documents should be pre-split into sentences. In more detail, the primary inputs are the following three items, each represented in JSON:  

\begin{itemize}
\item \textbf{conversation}: A list of conversational turns between the user and the assistant, where each item in the list is a dictionary with fields \verb+role+ and \verb+content+. The \verb+role+ equals to either \verb+user+ or \verb+assistant+, denoting user and assistant turns, respectively, while the \verb+content+ field contains the corresponding user/assistant utterance. The conversation should end with an assistant turn and the \verb+text+ field of that turn should contain the assistant utterance with each sentence prefixed with a response sentence ID of the form \verb+<rI>+, where \verb+I+ is an integer. The numbering should start from 0 (for the first sentence) and be incremented by one for each subsequent sentence in the last assistant turn. Note that only the last assistant turn should be split into sentences as described above; earlier assistant turns (as well as all user turns) should be maintained in their original form.
\item \textbf{instruction}: A task instruction, which is encoded as a dictionary with fields \verb+role+ and \verb+content+, where \verb+role+ equals to \verb+system+ and \verb+content+ equals to the following string describing the citation generation task: "Split the last assistant response into individual sentences. For each sentence in the response, identify the statement IDs from the documents that it references. Ensure that your output includes all response sentence IDs, and for each response sentence ID, provide the corresponding referring document sentence IDs." 
\item \textbf{documents}: A list of documents, where each item in the list is a dictionary with fields \verb+doc_id+ and \verb+text+. The \verb+text+ field contains the text of the corresponding document with each sentence prefixed with a context sentence ID of the form \verb+<cI>+, where \verb+I+ is an integer. The context sentence ID numbers should start from 0 (for the first sentence of the first document) and be incremented by one for each subsequent sentence. The numbers should continue to be incremented across documents to ensure that each context sentence ID appears once across the entire list of documents. For instance, if the last sentence of the 1st document has context sentence ID \verb+<cn>+, then the first sentence of the 2nd document is expected to have ID \verb=<cn+1>=.  
\end{itemize}

To prompt the LoRA adapter, we combine the above components as follows: We first append the \verb+instruction+ to the end of the \verb+conversation+ to generate an \verb+augmented_conversation+ list. Then we invoke the \verb+apply_chat_template+ function with parameters: \verb+conversation = augmented_conversation+ and \verb+documents = documents+.

\textbf{Model output:} When prompted with the above input, the model generates the citations for each sentence of the last assistant response in the form of a JSON array. The array is of the form \verb+[{"r": 0, "c": [...]}, {"r": 1, "c": [...]}, ...}]+, where a JSON object of the form \verb+{"r": k, "c": [l, m]}+, where \verb+k, l, m+ integers, denotes that the response sentence with ID \verb+<rk>+ is supported by the context sentences with IDs \verb+<cl>+ and \verb+<cm>+.

\quickstartpointer

\subsection{Evaluation}

We evaluate the LoRA adapter on two citation benchmarks:

\begin{itemize}
\item \textbf{ALCE} \cite{gao-etal-2023-enabling}: Evaluates the ability of models to produce \emph{document/passage-level} citations (i.e., identify the documents/passages that support a statement in the response).\\
\item \textbf{LongBench-Cite} \cite{zhang2024longciteenablingllmsgenerate}: Evaluates the ability of models to produce fine-grained \emph{span-level} citations (i.e., identify the spans within the input documents/passages that support a statement in the response) with a focus on long contexts.
\end{itemize}

Since the LoRA adapter is a post-hoc citation generation approach, its performance on the two benchmarks depends on the assistant responses for which it is asked to generate citations. To facilitate an apples-to-apples comparison, for each experiment, we keep the assistant responses the same and change the model that is used to generate the citations. In particular, we prompt an LLM to create an assistant response together with citations and evaluate the generated citations on the corresponding benchmark. Then, we compute and evaluate the citations generated for the same LLM response by the LoRA adapter.    

\subsubsection{Evaluation on ALCE}
\label{sec:citation-eval-alce}

For the ALCE evaluation, we prompt Llama-3.1-70B-Instruct and Mixtral-8x22B-Instruct to generate both the assistant response and corresponding passage-level citations. We first calculate the performance of the citations generated by these models on ALCE. Subsequently, we feed the responses of these models (leaving out the citations) to the LoRA adapter and evaluate its generated citations. The results are shown in Table \ref{tbl:citation-eval-alce}.

\begin{table}[h!]
\centering
\begin{tabular}{llccc}
\toprule 
{\bf Model generating response} & {\bf Model generating citations} & {\bf Recall} & {\bf Precision} & {\bf F1}\\
\midrule 
Llama-3.1-70B-Instruct & Llama-3.1-70B-Instruct & 61.4 & 58.1 & 59.7\\
Llama-3.1-70B-Instruct & Granite-3.3-8B LoRA citations & 55.4 & 65.0 & 59.8\\
Mixtral-8x22B-Instruct & Mixtral-8x22B-Instruct & 62.2 & 62.5 & 62.3\\
Mixtral-8x22B-Instruct & Granite-3.3-8B LoRA citations & 55.6 &  69.0 & 61.6\\
\bottomrule 
\end{tabular}
\caption{Citation generation evaluation on ALCE}
\label{tbl:citation-eval-alce}
\end{table}

We observe that the LoRA adapter performs on par with much bigger models when those are prompted to create passage-level citations. It is interesting to note that while the adapter's F1 performance is similar to the baselines, it exhibits a different precision-recall trade-off, trading lower recall for higher precision.   

Notes:
\begin{itemize}
\item All results are reported on the ELI5 dataset using the ORACLE (5-psg) setting.
\item To prompt Llama and Mixtral, we employ a setting similar to the one proposed in the ALCE paper; in particular we use a two-shot prompt comprised of two of the ICL examples from ALCE as well as a slightly modified version of the instruction from the paper \cite{gao-etal-2023-enabling}.
\item Sentence splitting of context/response is performed using NLTK.
\item Finally, since ALCE expects passage-level citations, we elevate the finer-grained citations produced by the LoRA adapter to the passage level before running the ALCE evaluation.
\end{itemize}

\subsubsection{Evaluation on LongBench-Cite}
\label{sec:citation-eval-longbenchcite}

For the LonBench-Cite evaluation, we prompt Llama-3.1-70B-Instruct to generate both the assistant response and corresponding citations. Then we evaluate the citations generated by Llama as well as the post-hoc citations generated by the LoRA adapter when invoked on the Llama responses. The results are shown in Table \ref{tbl:citation-eval-longbenchcite}.

\begin{table}[h!]
\centering
\small
\begin{tabular}
{p{1.6cm}|p{1.6cm}|>{\centering\arraybackslash}p{0.4cm}|>{\centering\arraybackslash}p{0.4cm}|>{\centering\arraybackslash}p{0.4cm}|>{\centering\arraybackslash}p{0.4cm}|>{\centering\arraybackslash}p{0.4cm}|>{\centering\arraybackslash}p{0.4cm}|>{\centering\arraybackslash}p{0.4cm}|>{\centering\arraybackslash}p{0.4cm}|>{\centering\arraybackslash}p{0.4cm}|>{\centering\arraybackslash}p{0.4cm}|>{\centering\arraybackslash}p{0.4cm}|>{\centering\arraybackslash}p{0.4cm}|>{\centering\arraybackslash}p{0.4cm}}
\toprule 
\multirow{2}{1.6cm}{\bf Model generating response} & \multirow{2}{1.6cm}{\bf Model generating citations} & \multicolumn{3}{p{1.6cm}|}{\bf Longbench-Chat (en)} & \multicolumn{3}{p{1.2cm}|}{\bf MultifieldQA (en)} & \multicolumn{3}{p{1.2cm}|}{\bf HotpotQA} & \multicolumn{3}{p{1.2cm}|}{\bf GovReport} & {\bf AVG}\\
\cmidrule{3-15} 
&  & {\bf R} & {\bf P} & {\bf F1} & {\bf R} & {\bf P} & {\bf F1} & {\bf R} & {\bf P} & {\bf F1} & {\bf R} & {\bf P} & {\bf F1} & {\bf F1}\\
\midrule 
Llama-3.1-70B-Instruct & Llama-3.1-70B-Instruct & 27.0 & 34.4 & 26.1 & 46.1 & 63.3 & 49.7 & 34.0 & 39.4 & 30.2 & 55.0 & 77.5 & 62.0 & 42.0\\
\midrule 
Llama-3.1-70B-Instruct & Granite-3.3-8B LoRA citations & 57.6 &  60.3 & 58.4 & 71.5 & 82.5 & 75.0 & 65.3 & 71.3 & 63.8 & 72.8 & 83.5 & 77.2 & 68.6\\
\bottomrule 
\end{tabular}
\caption{Citation generation evaluation on LongBench-Cite}
\label{tbl:citation-eval-longbenchcite}
\end{table}

We observe that the LoRA adapter performs across the board significantly better than Llama-3.1-70B-Instruct when prompted to create span-level citations. This demonstrates the value of the adapter to create post-hoc citations even for assistant responses generated by much bigger LLMs.

Notes:
\begin{itemize}
\item The evaluation results are reported on the English subset of LongBench-Cite (i.e., restricted to instances whose \verb+language+ field equals to \verb+en+).
\item In addition to the evaluation results per dataset, we also report the AVG F1, computed as the average of the four dataset-specific F1 scores.
\item The results for the LoRA adapter do not include the performance for 4/585 tasks, which encountered out of memory errors. 
\item To prompt Llama to generate a response with citations, we use the one-shot prompt described in the LongBench-Cite paper \cite{zhang2024longciteenablingllmsgenerate}.
\item For the LoRA adapter, sentence splitting of the context is performed using NLTK. For the response, we reuse the splitting in Llama's output (since the LongBench-Cite prompt instructs the model to output a response split into sentences/statements).
\end{itemize}

\subsection{Training Details}

The LoRA adapter was trained on synthetically-generated citation datasets. The process of generating the training data consisted of two main steps:

\begin{itemize}
\item \textbf{Multi-turn RAG conversation generation:} Starting from publicly available document corpora, we generated a set of multi-turn RAG data, consisting of multi-turn conversations grounded on passages retrieved from the corpora. For details on the RAG conversation generation process please refer to the Granite Technical Report\footnote{https://github.com/ibm-granite/granite-3.0-language-models/blob/main/paper.pdf} as well as \cite{lee2024multidocumentgroundedmultiturnsynthetic}.
\item \textbf{Citation generation:} For each turn of the multi-turn RAG conversations from the previous step, we used a multi-step synthetic citation generation pipeline to generate citations for the assistant response.
\end{itemize}

The following public datasets were used as seed datasets for the multi-turn RAG conversation generation process:

\renewcommand{\UrlFont}{\ttfamily\tiny}

\begin{itemize}
\item CoQA Wikipedia Passages (\url{https://stanfordnlp.github.io/coqa/})
\item MultiDoc2Dial (\url{https://huggingface.co/datasets/IBM/multidoc2dial})
\item QuAC (\url{https://huggingface.co/datasets/allenai/quac})
\end{itemize}

\renewcommand{\UrlFont}{\ttfamily\normalsize}

Leveraging the generated training data, the LoRA adapter was fine-tuned using PEFT under the following regime: rank = 8, learning rate = 1e-5, and 90/10 split between training and validation.

%% file: composite_qr_ad.tex
\subsection{Query Rewrite plus Answerability Determination}
\label{composite_query_rewrite_answerability}

\begin{figure*}[t]
    \centering
    \includegraphics[width=\textwidth]{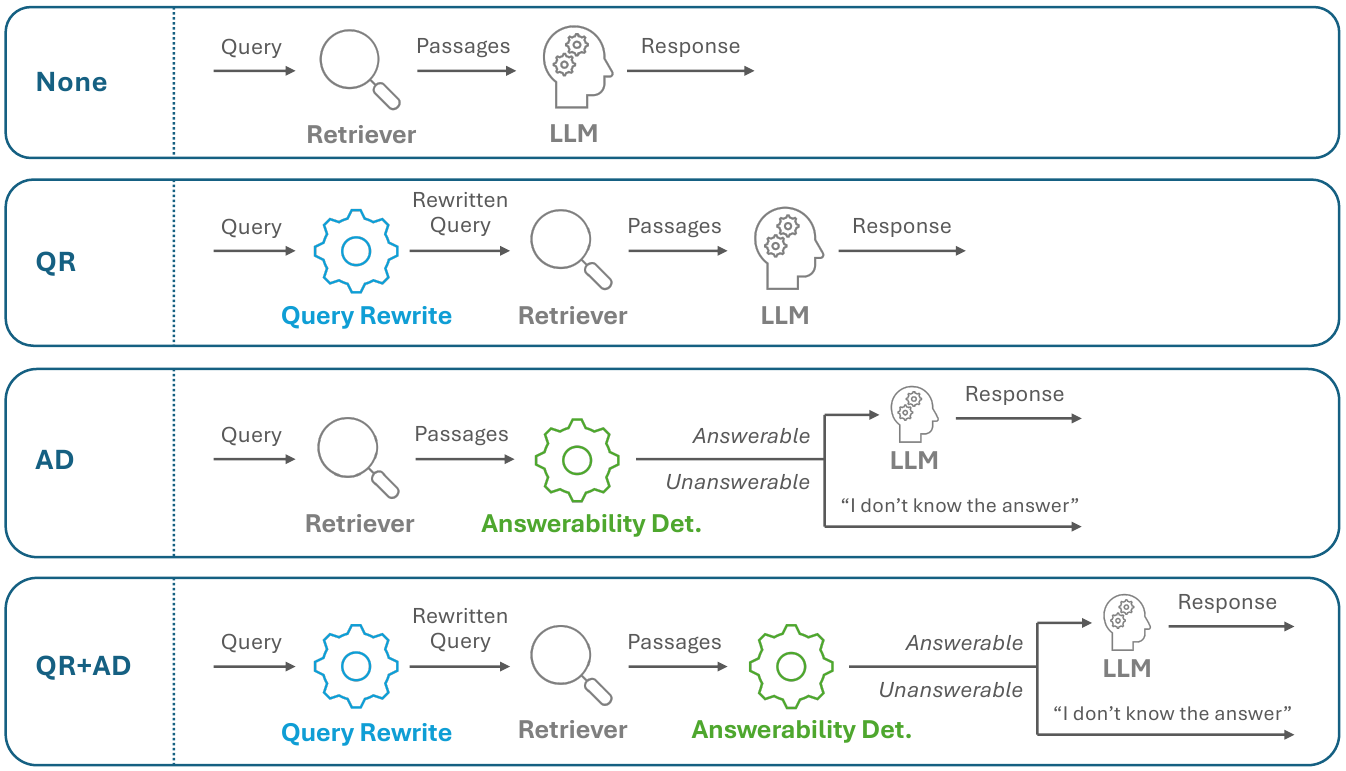}
\caption{RAG flows considered in this work}
\label{fig:rag-flows}
\end{figure*}

We briefly introduce four RAG flows which use these two intrinsics in various ways:\footnote{These are not the only possible applications of either intrinsic; rather these serve as reasonable examples that allow us to investigate the effects of composing them.}

\begin{itemize}
    \item{\textbf{None}: The given user query is used to retrieve the top $k$ passages; both are input to the generator model to create the response.}
    \item{\textbf{Query Rewrite (QR)}: The given user query is transformed using the QR instrinsic. The resulting query is used to retrieve the top $k$ passages. The original query and retrieved passages are input to the generator model to create the response.}
    \item{\textbf{Answerability Determination (AD)}: The given user query is used to retrieve the top $k$ passages. The query and passages are input to the AD intrinsic. If AD returns \textit{yes}, the query and passages are input to the generator model to create the response; if AD returns \textit{no}, this step is skipped and a pre-determined response of "I don't know the answer" is output.}
    \item{\textbf{Query Rewrite and Answerability Determination (QR+AD)}: Both intrinsics are used: QR to affect which top $k$ passages are retrieved and AD to determine whether to circumvent the generator model.}
\end{itemize}

Figure~\ref{fig:rag-flows} offers a visual representation of these four flows. We will now examine the benefits and tradeoffs to using both of these intrinsics in the four flows. As previously mentioned, the value of QR is in improving the performance of the retriever, increasing the relevance of the top $k$ passages. This increases the ability to generate both a more faithful response and more accurate citations, since there is more relevant context provided. On the other hand, the value of AD is in restraining an overeager model when a grounded response is impossible, greatly increasing the likelihood of correctly refusing to answer (though at the cost of occasionally being too conservative). Therefore, we must look at the quantitative effect of the individual and composite intrinsic flows on a) correctly classifying the query as answerable or unanswerable; b) the faithfulness of those responses which the generative model creates; and c) the aggregate score of faithfulness weighted by answerability classification. We will take these one at a time.

\textbf{Experimental Setup.} To benchmark the above flows, we use MT-RAG conversations and Elser for retriever. QR and AD is performed with QR LoRA adapter and AD LoRA adapter with Granite 3.3-8B Instruct, respectively. We set the number of retrieved passages to $5$ across different retrieval strategies. For generation, we send the Granite 3.3-8B Instruct model the following information: the entire conversation, top-$5$ retrieved passages, and the model's default RAG instruction prompt. 

\subsubsection{Evaluation: Answerability Classification}

Table \ref{tbl:qr_ad_answerability_classification} shows the $F1$ scores on the task of answerability classification of the 4 RAG flows described above. 
Using the AD intrinsic significantly improves performance on unanswerable queries ($F1$ score increases from 17 to 69). The QR intrinsic also has an effect (though it's much smaller): increasing the Recall@5 performance of the retriever makes some questions more likely to be answerable ($F1$ score increases from 77 to 83).

\begin{table}
    \centering
    \begin{tabular}{rcc}
    \toprule
     {\bf Flow}    & {\bf $F1_{Unanswerable}$} & {\bf $F1_{Answerable}$} \\
     \midrule
     None    & 17 & 77 \\
     QR    & 10 & 83 \\
     AD    & 69 & 81 \\
     QR+AD    & 51 & 82 \\
    \bottomrule
    \end{tabular}
  \caption{Performance on the task of Answerability Classification}
  \label{tbl:qr_ad_answerability_classification}
\end{table}

\subsubsection{Evaluation: Answer Faithfulness}

The QR intrinsic improves the faithfulness of the answer created by the generative model. Table \ref{tbl:qr_ad_answerables_ragasf} shows the number of responses in each of the 4 flows (meaning, the question was correctly identified as answerable and the generative model wrote a response) along with the RAGAS-F score (see Section~\ref{sec:query_rewrite}) on those responses. To identify whether a flow considers a query to have been answerable, a simple "I don't know" judge is used on the final output response, which determined whether the response contains content, or is in essence equivalent to saying, "I don't know the answer" (see \cite{katsis2025mtrag} for details on the IDK judge). It is important to note that this does not provide a comprehensive view, as it does not reflect the performance of the RAG system on the rest of the cases not captured by this table (thus the inclusion of the number of responses that was able to be scored in each flow).

\begin{table}
    \centering
    \begin{tabular}{rcc}
    \toprule
     {\bf Flow}    & {\bf \#Responses} & {\bf RAGAS-F} \\
     \midrule
     None    & 512 & 74 \\
     QR    & 585 & 76 \\
     AD    & 427 & 66 \\
     QR+AD    & 506 & 68 \\
    \bottomrule
    \end{tabular}
  \caption{Faithfulness of generated responses (RAGAS-F) for queries determined to be answerable by each flow; the size of that set is denoted by \#Responses (Number of Generated Responses)}
  \label{tbl:qr_ad_answerables_ragasf}
\end{table}

\subsubsection{Evaluation: Joint Answerability-Faithfulness}

Considering the RAGAS-F score from Table \ref{tbl:qr_ad_answerables_ragasf} in isolation, it would appear that using the AD intrinsic harms performance, and that the best approach is to only make use of the QR intrinsic. Therefore it is clear that we should not only rely on this evaluation. Therefore, we return to the Joint Answerability-Faithfulness Score (JAFS), introduced in Section~\ref{sec:answerability}. This score rewards the model for correctly abstaining on unanswerable queries (full credit) and for providing faithful answers on answerable queries (partial credit based on RAGAS Faithfulness). No credit is given for responding to an unanswerable query, nor for refusing to respond to an answerable query. Table~\ref{tbl:qr_ad_joint_a_f} presents the JAFS score for each of the four flows. 

\begin{table}
    \centering
    \begin{tabular}{rc}
    \toprule
     {\bf Flow}    & {\bf JAFS}  \\
     \midrule
     None    & 49  \\
     QR    & 56  \\
     AD    & 66 \\
     QR+AD    & 66  \\
    \bottomrule
    \end{tabular}
\caption{Joint Answerability-Faithfulness Score for each flow}
  \label{tbl:qr_ad_joint_a_f}
\end{table}

By going through this careful evaluation process, we are able to understand the benefits and trade-offs of this composite flow. The important aspects were a) creating appropriate non-composite flows for comparison and b) analyzing the effect of each flow using metrics which reflect the value brought by each intrinsic. We have now been able to demonstrate that for the use case where the conversations with our RAG system are expected to frequently be multi-turn, and it is important to limit responses to only those which can be successfully supported, making use of both the QR and AD intrinsic in the flow as described will yield better overall performance.

%% file: appendices.tex
\subsection{Changes in performance from Granite Instruct 3.2 to 3.3}

In this section we compare performance for intrinsics fine-tuned on Granite 3.3 8b Instruct to earlier versions trained on Granite 3.2 8b Instruct (we found no significant difference in performance for the Query Rewrite, Hallucination Detection, and Uncertainty Quantification intrinsics.)

\subsubsection{Answerability Determination}

We compare the performance of the answerability determination intrinsic fine-tuned on Granite 3.3 8b Instruct to an earlier version trained on Granite 3.2 8b Instruct. Table~\ref{tbl:answerability-eval-granite-compare-mtrag} and Table~\ref{tbl:answerability-eval-granite-compare-squad} presents a detailed evaluation on the MT-RAG and SQUAD datasets respectively, reporting precision, recall, and F1 scores for both answerable and unanswerable queries. Compared to Granite 3.2, the Granite 3.3-based model shows improved recall on answerable examples, but this comes at the expense of performance on unanswerable queries. Aggregate performance remains quite comparable across models, though Granite 3.3 performs slightly worse on SQUAD.

\begin{table}[h!]
\centering
\small
\begin{tabular}{cccccccc}
\toprule 
& \multicolumn{3}{c}{unanswerable} & \multicolumn{3}{c}{answerable} & \multirow{2}{*}{W.F1} \\
  & P & R & F1 & P & R & F1 &  \\
 \midrule 
\begin{tabular}[c]{@{}c@{}} Granite 3.2 8b \\ Instruct LoRA\end{tabular} & 89.1 & 93.7 & 91.3 & 92.3 & 86.7 & 89.4 & 90.5 \\
\midrule
Granite   3.3 8b \\ Instruct LoRA & 89.8 & 91.8 & 90.8 & 90.3 & 87.9 & 89.1 & 90 \\
\bottomrule 
\end{tabular}
\caption{Evaluation of the Granite 3.2 and 3.3 versions of the answerability classification intrinsic on MT-RAG.}
\label{tbl:answerability-eval-granite-compare-mtrag}
\end{table}

\begin{table}[h!]
\centering
\small
\begin{tabular}{ccccccccccccccc}
\toprule 
 & \multicolumn{3}{c}{unanswerable} & \multicolumn{3}{c}{answerable} & \multirow{2}{*}{W.F1} \\
 & P & R & F1 & P & R & F1 &  \\
 \midrule 
\begin{tabular}[c]{@{}c@{}} Granite 3.2 8b \\ Instruct LoRA\end{tabular} & 84.2 & 68 & 75.2 & 73.1 & 87.2 & 79.5 & 77.4 \\
\midrule
Granite   3.3 8b \\ Instruct LoRA & 88.1 & 59.3 & 70.9 & 69.3 & 92 & 79 & 75 \\
\bottomrule 
\end{tabular}
\caption{Evaluation of the Granite 3.2 and 3.3 versions of the answerability classification intrinsic on SQUAD.}
\label{tbl:answerability-eval-granite-compare-squad}
\end{table}

\subsubsection{Citation Generation}

We compare the performance of the latest citation generation intrinsic trained on IBM Granite 3.3 8b instruct to an earlier version of the intrinsic trained on the IBM Granite 3.2 8b instruct model. Tables \ref{tbl:citation-eval-alce-extended} and \ref{tbl:citation-eval-longbenchcite-extended} show the performance of the two versions of the intrinsic compared to the baselines when evaluated on ALCE and LongBench-Cite, respectively. The experimental settings are identical to the ones described in Sections \ref{sec:citation-eval-alce} and \ref{sec:citation-eval-longbenchcite} for ALCE and LongBench-Cite, respectively.

\begin{table}[h!]
\centering
\begin{tabular}{llccc}
\toprule 
{\bf Model generating response} & {\bf Model generating citations} & {\bf Recall} & {\bf Precision} & {\bf F1}\\
\midrule 
Llama-3.1-70B-Instruct & Llama-3.1-70B-Instruct & 61.4 & 58.1 & 59.7\\
Llama-3.1-70B-Instruct & Granite-3.2-8B LoRA citations & 54.8 & 65.9 & 59.8\\
Llama-3.1-70B-Instruct & Granite-3.3-8B LoRA citations & 55.4 & 65.0 & 59.8\\
Mixtral-8x22B-Instruct & Mixtral-8x22B-Instruct & 62.2 & 62.5 & 62.3\\
Mixtral-8x22B-Instruct & Granite-3.2-8B LoRA citations & 54.3 & 69.5 & 61.0\\
Mixtral-8x22B-Instruct & Granite-3.3-8B LoRA citations & 55.6 &  69.0 & 61.6\\
\bottomrule 
\end{tabular}
\caption{Evaluation of the Granite 3.2 and 3.3 versions of the citation generation intrinsic on ALCE}
\label{tbl:citation-eval-alce-extended}
\end{table}

\begin{table}[h!]
\centering
\small
\begin{tabular}
{p{1.6cm}|p{1.6cm}|>{\centering\arraybackslash}p{0.4cm}|>{\centering\arraybackslash}p{0.4cm}|>{\centering\arraybackslash}p{0.4cm}|>{\centering\arraybackslash}p{0.4cm}|>{\centering\arraybackslash}p{0.4cm}|>{\centering\arraybackslash}p{0.4cm}|>{\centering\arraybackslash}p{0.4cm}|>{\centering\arraybackslash}p{0.4cm}|>{\centering\arraybackslash}p{0.4cm}|>{\centering\arraybackslash}p{0.4cm}|>{\centering\arraybackslash}p{0.4cm}|>{\centering\arraybackslash}p{0.4cm}|>{\centering\arraybackslash}p{0.4cm}}
\toprule 
\multirow{2}{1.6cm}{\bf Model generating response} & \multirow{2}{1.6cm}{\bf Model generating citations} & \multicolumn{3}{p{1.6cm}|}{\bf Longbench-Chat (en)} & \multicolumn{3}{p{1.2cm}|}{\bf MultifieldQA (en)} & \multicolumn{3}{p{1.2cm}|}{\bf HotpotQA} & \multicolumn{3}{p{1.2cm}|}{\bf GovReport} & {\bf AVG}\\
\cmidrule{3-15} 
&  & {\bf R} & {\bf P} & {\bf F1} & {\bf R} & {\bf P} & {\bf F1} & {\bf R} & {\bf P} & {\bf F1} & {\bf R} & {\bf P} & {\bf F1} & {\bf F1}\\
\midrule 
Llama-3.1-70B-Instruct & Llama-3.1-70B-Instruct & 27.0 & 34.4 & 26.1 & 46.1 & 63.3 & 49.7 & 34.0 & 39.4 & 30.2 & 55.0 & 77.5 & 62.0 & 42.0\\
\midrule 
Llama-3.1-70B-Instruct & Granite-3.2-8B LoRA citations & 61.9 & 68.6 & 62.0 & 71.2 & 84.1 & 74.3 & 66.8 & 73.3 & 65.4 & 70.3 & 83.6 & 75.4 & 69.3\\
\midrule 
Llama-3.1-70B-Instruct & Granite-3.3-8B LoRA citations & 57.6 &  60.3 & 58.4 & 71.5 & 82.5 & 75.0 & 65.3 & 71.3 & 63.8 & 72.8 & 83.5 & 77.2 & 68.6\\
\bottomrule 
\end{tabular}
\caption{Evaluation of the Granite 3.2 and 3.3 versions of the citation generation intrinsic on LongBench-Cite}
\label{tbl:citation-eval-longbenchcite-extended}
\end{table}

We observe that the two versions of the citations generation intrinsic exhibit similar performance. When evaluated on ALCE, the intrinsic trained on IBM Granite 3.3 8b instruct performs the same or slightly better  than the one trained on 3.2, while when evaluated on LongBench-Cite, the 3.3 version is overall slightly worse than the 3.2 version. However, the difference is pretty small with the overall performance of the two versions being in all conducted experiments within 0.7 percentage points of each other.